\documentclass[sigconf]{acmart}
\makeatletter
\renewcommand{\maketag@@@}[1]{\hbox{\m@th\normalsize\normalfont#1}}%
\makeatother

\usepackage{microtype}
\usepackage{makecell}
\usepackage{graphicx}
\usepackage{subcaption}
\usepackage{booktabs}
\usepackage{amsmath}
\usepackage{mathtools}
\usepackage{amsthm}
\usepackage{multirow}
\usepackage{makecell}
\usepackage{float}
\usepackage{caption}
\usepackage{comment}
\usepackage{algorithm}
\usepackage{algorithmic}
\usepackage{graphicx}
\usepackage{booktabs}
\usepackage{multirow}
\usepackage{subcaption}
\usepackage{pgfplots}
\pgfplotsset{compat=1.18}

\setcopyright{acmlicensed}
\copyrightyear{2026}
\acmYear{2026}
\acmDOI{10.1145/3767308.3838608}

\setcopyright{cc}
\setcctype{by}
\acmConference[MM '26] {Proceedings of the 34th ACM International Conference on Multimedia}{November 10--14, 2026}{Rio de Janeiro, Brazil.}
\acmBooktitle{Proceedings of the 34th ACM International Conference on Multimedia (MM '26), November 10--14, 2026, Rio de Janeiro, Brazil}
\acmISBN{979-8-4007-2213-4/2026/11}

\begin{document}

\title{ Explainable Deepfake Detection with Feature-robust Augmentation and Evidence-grounded Explanation Optimization}


\author{Zhu Xu}
\affiliation{%
  \institution{Wangxuan Institute of Computer\\ Technology, Peking University}
  \city{Beijing}
  \country{China}}
\email{xuzhu@stu.pku.edu.cn}

\author{Jiaqi Tang}
\affiliation{%
  \institution{Wangxuan Institute of Computer\\ Technology, Peking University}
 \city{Beijing}
  \country{China}}
\email{2601213445@stu.pku.edu.cn }

\author{Pokai Chen}
\affiliation{%
  \institution{Wangxuan Institute of Computer\\ Technology, Peking University}
 \city{Beijing}
  \country{China}}
\email{2501213417@stu.pku.edu.cn }

\author{Yuxin Peng}
\affiliation{%
  \institution{Wangxuan Institute of Computer\\ Technology, Peking University}
  \city{Beijing}
  \country{China}}
\email{pengyuxin@pku.edu.cn}

\author{Yang Liu}
\authornote{Corresponding author}
\affiliation{%
  \institution{Wangxuan Institute of Computer\\ Technology, Peking University}
  \city{Beijing}
  \country{China}}
\email{yangliu@pku.edu.cn}

\renewcommand{\shortauthors}{Zhu Xu, Jiaqi Tang, Bokai Chen, Yuxin Peng, \& Yang Liu}

\renewcommand{\shortauthors}{}

\begin{abstract}
Explainable deepfake detection extends binary classification by requiring models to not only predict authenticity but also provide interpretable justifications. This expanded scope is critical in practice, where users like forensic analysts need insight into the rationale behind the detection.
Despite advancements, current approaches suffer from two critical deficiencies: (1)\textit{vulnerability to image quality degradation}: detection accuracy plummets on low-quality samples, while naive augmentation strategies may induce feature drift and impair performance as diversity expands. (2) \textit{factually flawed explanations}: explanation models may omit manipulation evidence or hallucinate irrelevant details, undermining interpretability. To address it, we propose a framework with two innovations. For robust deepfake detection, we introduce \textit{Feature-robust Augmentation}, which comprises diversified degradation-aware augmentation strategies, and a supervised contrastive learning pattern paired with a mean-teacher architecture that stabilizes features against augmentations through consistency constraints. For explanation, we devise an \textit{evidence-grounded preference optimization} process that guides model to prioritize genuine manipulation traces by learning from chosen-rejected explanation pairs, where rejected samples are constructed via evidence omission or irrelevant information injection. The proposed approach wins the first place in ACM
Multimedia 2026 Explainable Deepfake Detection Challenge.
The code is available at \href{https://github.com/oceanflowlab/EDD.git}{https://github.com/oceanflowlab/EDD.git}

\end{abstract}



\maketitle
\section{Introduction}

The rapid advancement of generative AI has made image manipulation increasingly realistic and challenging to verify, posing serious risks to digital media integrity and forensic investigations~\citep{esser2024scaling, tian2024visual, huang2024ffaa, guo2025rethinking, peng2025mllm, ren2025can, tariq2025llms}. Although the majority of existing deepfake detection formulates this as a binary decision task, i.e., classifying an image as real or fake, such formulation fails to capture the practical demands. In practice, a label alone is seldom sufficient: fact-checkers and forensic experts must comprehend the specific visual evidence that renders an image suspect. Consequently, the field has increasingly recognized the necessity of explainable detection, wherein systems are expected to both make accurate predictions and articulate the visual cues underpinning those decisions, e.g., inconsistent object boundaries, implausible geometry, etc. By providing human-interpretable reasoning, such systems can facilitate more effective forensic scrutiny and foster greater user confidence in automated detection tools.



\begin{figure}[thbp]
    \centering
    \begin{subfigure}[b]{0.46\textwidth}
        \centering
        \includegraphics[width=\textwidth]{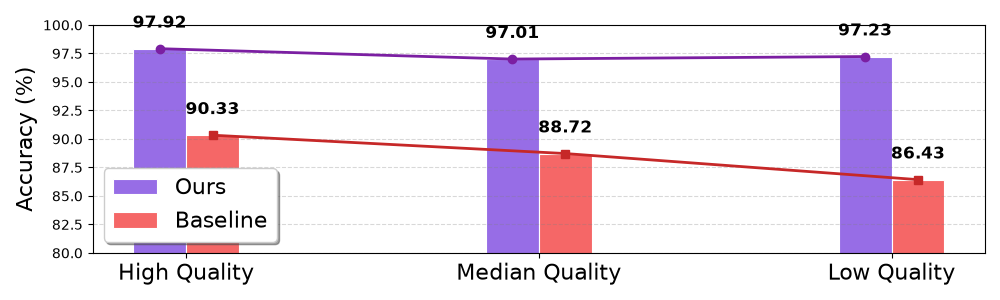}
        \vspace{-2mm}
        \caption{Detection accuracy on samples with different image quality.}
        \label{fig:exp1}
    \end{subfigure}
    \hfill
    \begin{subfigure}[b]{0.47\textwidth}
        \centering
        \includegraphics[width=\textwidth]{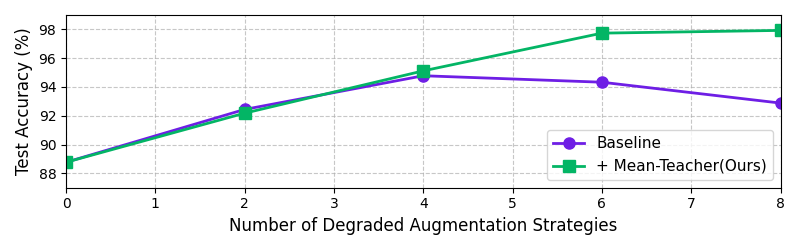}
        \vspace{-2mm}
        \caption{Performance change under increasing augmentations.}
        \label{fig:exp2}
    \end{subfigure}
    \vspace{-3mm}
    \caption{Illustration of analysis for detection robustness. (a) Performance drops under quality degradation. (b) Mean-teacher stabilizes learning under diverse augmentations while the baseline suffers from feature drift.}
    \label{fig:overall}
\end{figure}

Despite advancements~\citep{kuckreja2026pixelsdontliebut,huang2024ffaa, guo2025rethinking, peng2025mllm} in explainable deepfake detection, existing methods face two fundamental challenges that severely undermine their real-world applicability. (1)\textit{the brittleness of detection models under image quality degradation.} 
As shown in Figure~\ref{fig:exp1}, we adopt BRISQUE\cite{brisque2024} as the metric to measure the image quality of the validation set of XPlainVerse Challenge dataset, partitioning the samples into three subsets with high, medium and low image quality. We then evaluate detection performance on them.
Performance on baseline(DINOv3\cite{siméoni2025dinov3} + classification head trained without any data augmentation, red bar) shows a decrease as the image quality drops, which confirms that quality degradation indeed compromises discriminative capability. A straightforward remedy, i.e., applying data augmentations, often backfires due to feature drift, where the model struggles to learn a consistent representation across augmented views. In Figure~\ref{fig:exp2}, as the number of augmentation strategies for training increases, the test accuracy of baselines(red color) initially rises but subsequently declines, indicating that excessively diverse augmentations blur classification boundaries and confuse the model.
(2)\textit{inadequate evidence-verification of existing explanation mechanisms.} Prevailing approaches predominantly focus on the procedural format of reasoning, i.e., mandating multi-step thinking or sufficient response length, while neglecting the factual accuracy and completeness of the evidence presented within the rationale. We evaluate explanations of the baseline Qwen3-VL-8B-Instruct\cite{yang2025qwen3} on a 1k-subset of validation set, observing that two prevalent failure modes emerge: (i) \textit{evidence omission}, where crucial manipulation traces are missing from the explanation, which exist in 68.3\% samples, and (ii) \textit{irrelevant information intrusion}, where spurious details are hallucinated that exist in 57.7\% samples.  These deficiencies not only mislead users but also erode the trust that explainability aims to establish.



To address these challenges, we propose a two-stage framework with two components. For detection robustness, we introduce a \textit{Feature-robust
augmentation} strategy, which first uses a degradation-aware augmentation pipeline that augments input images with diverse distortions at varying intensities. To complement this, we impose a supervised contrastive loss that pulls features of the same authenticity class(real or fake) into more compact clusters, enhancing discriminability across quality variations. However, contrastive learning operates only at the class level without preventing intra-instance feature drift. We therefore introduce a mean-teacher architecture that maintains a stable feature anchor via exponential moving average of the student's weights. The teacher aggregates information across consecutive training steps, suppressing noise and providing a consistent reference. A consistency constraint aligns all augmented student features toward this anchor, preventing intra-instance drift while preserving the benefits of augmentation diversity.
For deepfake explanation, we devise an \textit{Evidence-grounded Explanation Optimization} to guide the model via reinforcement learning. By constructing chosen-rejected explanation pairs where chosen explanations are complete and accurate, while rejected ones omit key evidence or include fabrications, we fine-tune the model to prioritize genuine manipulation evidence over superficial reasoning patterns via direct preference optimization(DPO)\cite{rafailov2023dpo}, enhancing the evidence accuracy and completeness of explanations. 

Beyond detailed evidence generation, we further develop a concise explanation model tailored for scenarios where brevity is prioritized. This model is optimized via GRPO\cite{shao2024grpo} algorithm, with rewards jointly designed on semantic fidelity and conciseness, enabling it to deliver only the most critical manipulation cues. Combined with the detailed evidence explanation model, our framework flexibly adapts to diverse usage scenarios from in-depth forensic analysis to quick public fact-checking. Evaluated on the 200k XPlainVerse\cite{narang2026xplainversemillionscalebenchmarkexplainable} test set, our model achieves state-of-the-art overall performance in ACM MM 2026 Explainable Deepfake Detection Challenge\cite{narang2026explainabledeepfakedetectionchallenge}. Specifically, we achieve the best semantic fidelity for evidence-grounded explanation, the optimal semantic fidelity and conciseness balance for concise explanation.

The contributions of this paper are summarized as follows:
\begin{itemize}
    \item We identify and empirically validate two critical vulnerabilities in existing explainable deepfake detection systems, i.e., sensitivity to image quality and lack of evidence accuracy verification.
    \item  We propose a framework that jointly enhances detection robustness via degradation-aware augmentation with mean-teacher stabilization, and improves explanation faithfulness through evidence-grounded preference optimization.
    \item Our framework shows state-of-the-art overall performance, ranks first for ACM MM 2026 Explainable Deepfake Detection Challenge\cite{narang2026explainabledeepfakedetectionchallenge}, offers a reliable solution for explainable deepfake detection.
\end{itemize}

\section{Related Work}
\label{sec:related}

\subsection{Robust Deepfake Detection}
Recent advances have explored various strategies to improve robustness against unseen forgeries. To diminish the impact of image quality on detector performance, several works~\cite{hopf2025practicalmanipulationmodelrobust,frank2020leveraging, qian2020thinking, tan2024frequency,zhou2024freqblender,liu2021spatialphaseshallowlearningrethinking,10891978} attempt to model different degradation strategies. Though promising, such naive data augmentation may lead to feature drift, causing classification boundary to blur as augmentation diversity expands.
Our method addresses such limitations by incorporating a mean-teacher architecture that maintains a stable feature anchor, enforcing consistency constraints across all augmented views to prevent feature drift while preserve the benefits of augmentation diversity. 

\subsection{Vision-Language Models for Explainable Deepfake Detection}
Recent advancements in Vision-Language Models (VLMs)~\citep{liu2023visual, bai2025qwen2, zhu2025internvl3} adopted a paradigm of supervised fine-tuning followed by reinforcement learning, which has proven effective in boosting reasoning capabilities across diverse applications, such as image\cite{wang2024cogvlmvisualexpertpretrained,chen2024internvl,11554560}, video\cite{zheng2026temporalawarereasoningoptimizationvideo,park2026deepvideor1videoreinforcementfinetuning} and 3D\cite{mo2026distillingneurosymbolicprograms3d,zhu2025llava3dsimpleeffectivepathway,Yang2026GaLa25DGA} understanding tasks. The rapid development of VLMs has also opened new possibilities for explainable deepfake detection.
For example, ForgerySleuth~\cite{sun2024forgerysleuth} employs a trace encoder to generate detailed tampering analyses. 
And more recent works\cite{tan2026veritasgeneralizabledeepfakedetection} propose datasets annotated by humans to generate explanations further aligned with human preference. However, the post-training focuses on the format of reasoning, i.e, requiring multi-step thinking or sufficient response length, while neglecting the factual accuracy and completeness of evidence. Different from them, we identify the critical issue of evidence inaccuracy, termed evidence omission and irrelevant information intrusion, and address it through an evidence-grounded preference optimization process. By constructing chosen-rejected explanation pairs, we fine-tune the model to prioritize genuine manipulation evidence over superficial reasoning patterns. 


\section{Method}

\subsection{Overview}

\begin{figure*}
    \centering
    \includegraphics[width=0.8\linewidth]{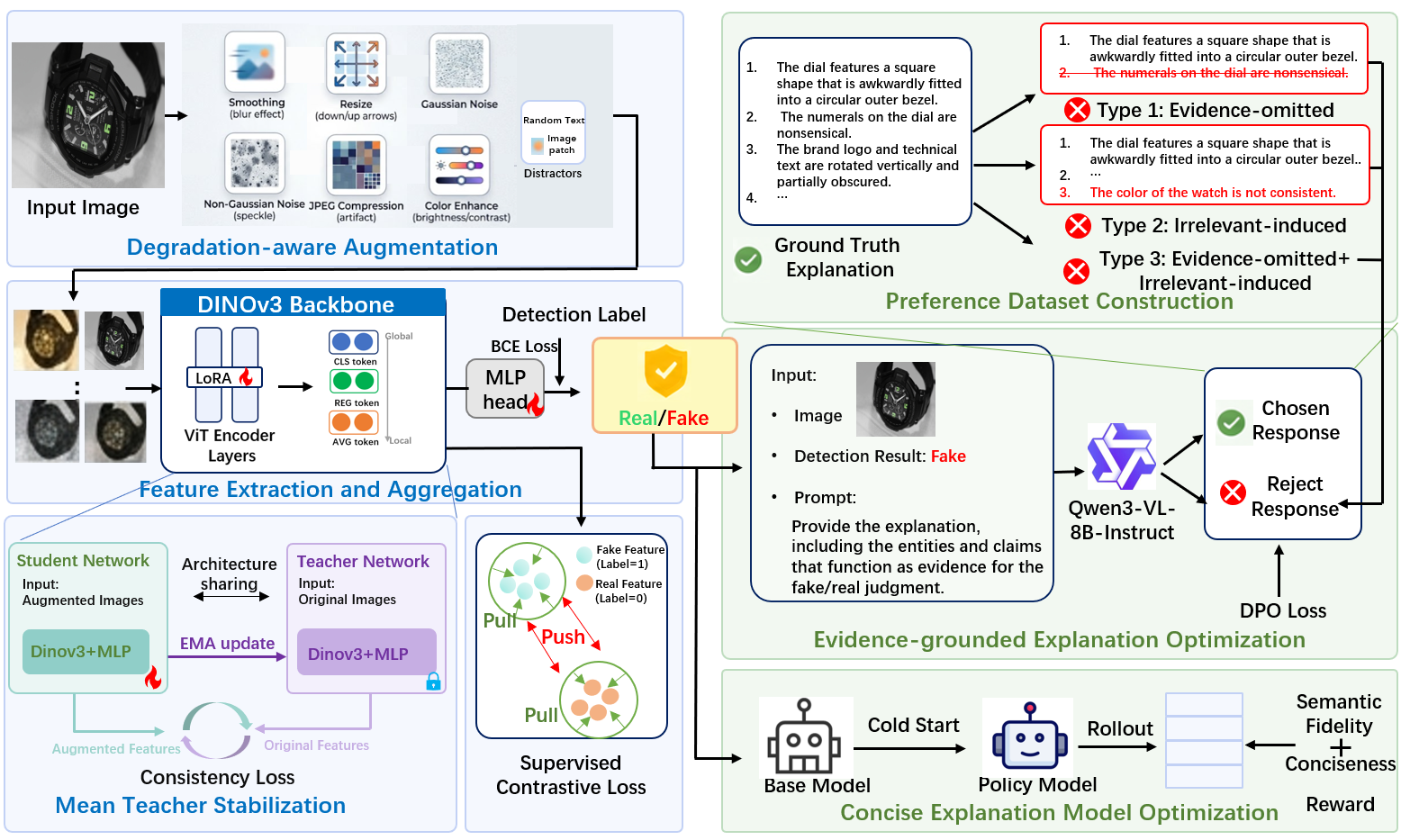}
    \vspace{-4mm}
    \caption{ \textbf{Overview of our proposed framework.} Given an input image, the visual backbone extracts features and fed into a classifier for prediction. During training, the detection branch incorporates degradation-aware augmentations, supervised contrastive learning, and mean-teacher consistency regularization to enhance robustness against quality degradation and feature drift.
    For explanation generation, the image and deepfake label are sent to the MLLM for generation. To enhance evidence accuracy and completeness, we propose an evidence-grounded optimization process, which generates evidence-omitted and irrelevant information induced explanation to construct rejected samples, guiding the model via the DPO algorithm. We also provide a concise explanation model that is trained via GRPO with tailored semantic fidelity and conciseness rewards. }
    \label{fig:framework}
\end{figure*}
Our goal is to build a deepfake detection system that is both robust against real-world quality variations and capable of providing faithful, evidence-grounded explanations. To this end, we propose a unified framework comprising two synergistic components, as illustrated in Figure~\ref{fig:framework}. The first component targets deepfake detection. Given an input image, the visual backbone extracts features and passes through a classifier for binary classification(1 for fake and 0 for real). To ensure robustness under diverse quality conditions, we incorporate a degradation-aware augmentation pipeline during training, paired with a mean-teacher architecture that anchors augmented representations toward a stable reference, preventing feature drift. We also incorporate a supervised contrastive loss to make features more distinguishable. Once a detection decision is made, the second component generates an explanation to justify its reasoning. For detailed forensic analysis, we employ an MLLM for the reasoning. We propose an evidence-grounded explanation optimization process, which explicitly teaches the model to prioritize genuine manipulation evidence over potential evidence omissions or hallucinations via DPO algorithm training, where we construct different sub-optimal responses to simulate potential problems in model reasoning. For scenarios where brevity is prioritized, we additionally provide a concise explanation model optimized via tailored reward functions. Together, these components deliver a detector that is not only accurate across diverse conditions but also transparent in its decision-making.


\subsection{Robust Deepfake Detection via Feature-robust Augmentation}


\subsubsection{Feature Extraction and Aggregation}

We employ DINOv3-7B~\cite{siméoni2025dinov3} as our visual encoder. To fully exploit both global and local information, we aggregate multi-granularity tokens. Specifically, we extract CLS token $\in \mathbb{R}^{1*d}$, REG token $\in \mathbb{R}^{4*d}$ with global semantics, AVG patch tokens $\in \mathbb{R}^{1*d}$ preserve fine-grained details, where $d$ is feature dimension. These tokens are concatenated and subsequently passed through an MLP head for classification, supervised by a binary cross-entropy loss $\mathcal{L}_{\rm BCE}$. This design ensures that the detector leverages both high-level contextual cues and fine-grained manipulation traces simultaneously.

\subsubsection{Degradation-Aware Augmentation Pipeline}
To build robustness against quality degradations, we design a augmentation pipeline that exposes model to diverse distortion types. Table~\ref{tab:degradations} summarizes our strategies and intensity sampling. 
Distractors (random text or images) are additionally applied to simulate real-world overlays. All degradations are applied in random probability and order, ensuring that the model encounters a wide spectrum of quality conditions during training.

\begin{table}[!t]
\centering
\caption{Summary of degradation strategies and intensities.}
\label{tab:degradations}
\vspace{-3mm}
\resizebox{0.47\textwidth}{!}{
\begin{tabular}{@{}l|c@{}}
\toprule
\textbf{Degradation} & \textbf{Sampling Intensities and Conditions} \\
\midrule
Smoothing & 
\makecell[c]{3 kernel types: anisotropic Gaussian ($\sigma \sim \mathcal{U}(0, 30)$),\\ fspecial Gaussian ($\sigma \sim \mathcal{U}(0, 30)$), uniform square ($w \sim \mathcal{U}(3, 30)$)} \\
\hline
Resize & \makecell[c]{Upsample $\sim \mathcal{U}(1,2)$, downsample $\sim \mathcal{U}(0.5,1)$,\\ interpolation: linear/cubic/area} \\
\hline
Gaussian Noise & \makecell[c]{3 types: per-channel, grayscale, correlated ;\\ two regimes: $\sigma \sim \mathcal{U}(2,100)$ and $\sigma \sim \mathcal{U}(80,100)$} \\
\hline
Non-Gaussian Noise & \makecell[c]{Speckle ($\sigma \sim \mathcal{U}(2,25)$)/ Poisson} \\ \hline
JPEG Compression & \makecell[c]{Quality factor $\sim \mathcal{U}(10, 95)$} \\ \hline
Color Enhance & \makecell[c]{Brightness or contrast factor $\sim \mathcal{U}(0.5, 1.5)$} \\ \hline
Distractors & \makecell[c]{Random text overlay or image patch overlay} \\
\bottomrule
\end{tabular}}
\end{table}
To enhance robustness against degradation variations and explicitly enforce the features more authenticity-aware, i.e., discriminative for real and fake prediction, we impose a supervised contrastive loss, as naively applying augmentations to expand training diversity can paradoxically induce feature drift, causing representations of augmented samples to deviate from their original distributions and impair discriminative performance. Supervised contrastive loss handles this by pulling features of the same authenticity class into compact clusters regardless of their degradation condition, while pushing features of different class apart. For a given image $x_i$ with label $y_i$, we generate a degraded view $x_i^d = t(x_i)$ where $t$ is sampled from our augmentation pipeline. The loss pulls features of the same class together while pushing features of different classes apart:
\vspace{-2mm}
\begin{equation}
\mathcal{L}_{\text{supcon}} = \sum_{i \in \mathcal{B}} \frac{-1}{|\mathcal{P}(i)|} \sum_{p \in \mathcal{P}(i)} \log \frac{\exp(\mathbf{x}_i \cdot \mathbf{x}_p / \tau)}{\sum_{a \in \mathcal{A}(i)} \exp(\mathbf{x}_i \cdot \mathbf{x}_a / \tau)},
\end{equation} 
where $\mathcal{B}$ is the current batch, $\mathcal{P}(i)$ denotes the set of positive samples sharing the same label as $i$ (including both original and degraded views), $\mathcal{A}(i)$ is the set of all anchors excluding $i$, and $\tau$ is a temperature parameter.

\subsubsection{Mean-Teacher Stabilization}
While supervised contrastive loss effectively enforces authenticity-aware feature clustering, it operates solely at the class level to ensure real and fake samples are more distinguishable without constraining how different augmented views of the same sample relate to each other beyond pulling them toward the same class centroid. As degradation diversity increases, this leaves room for intra-instance feature drift: different views of the same image may scatter widely within the same class cluster, undermining prediction stability and consistency. A potential remedy would be to impose a consistency loss between different augmented views of the same sample. However, this approach lacks a stable reference: when both views are independently perturbed, the model may simply enforce agreement between two noisy predictions without anchoring to a meaningful representation, causing features to drift arbitrarily. To complement this, we introduce a mean-teacher architecture that directly constrains each sample's feature across augmentations toward a stable reference. As proved by prior works\cite{tarvainen2018meanteachersbetterrole, moralesbrotons2024exponentialmovingaverageweights}, exponential moving average (EMA) aggregates information across consecutive model states, producing more accurate and consistent predictions by canceling out noise while preserving the signal \cite{laine2017temporalensemblingsemisupervisedlearning}. The teacher network $\theta_{\text{tea}}$, updated as an EMA of the student network $\theta_{\text{stu}}$, inherently suppresses stochastic degradation noises to provide stable sample features.
For each image $x_i$, we feed clean view and all degraded views into student network, while the teacher receives only clean view. A consistency loss enforces alignment between the student's features from degraded views and the teacher's feature from the clean view:
\vspace{-2mm}
\begin{equation}
\mathcal{L}_{\text{consis}} = \frac{1}{|\mathcal{T}|} \sum_{t \in \mathcal{T}} \| \mathbf{z}_\text{stu}(t(x_i)) - \mathbf{z}_\text{tea}(x_i) \|_2^2.
\end{equation}

where $\mathcal{T}$ is the set of all augmentations. This alignment constraint anchors all augmented representations toward a stable reference, preventing feature drift while preserving the benefits of augmentation diversity. The overall loss is:
\vspace{-2mm}
\begin{equation}
\mathcal{L}_{\text{det}} = \mathcal{L}_{\text{BCE}} + \lambda_1 \mathcal{L}_{\text{supcon}} + \lambda_2 \mathcal{L}_{\text{consis}}.
\end{equation}
where $\lambda_1$ and $\lambda_2$ are coefficients to balance the loss terms.




\subsection{ Evidence-Grounded Explanation Optimization}
Beyond accurate detection, our system is required to provide rationales that faithfully reflect the model's decision process. Our empirical findings reveal that existing explanation models frequently suffer from two critical issues: evidence omission, i.e., failing to mention crucial manipulation traces, and irrelevant information intrusion, i.e., hallucinating spurious details. While supervised fine-tuning (SFT) enables the MLLM to generate plausible explanations by mimicking ground-truth rationales, it does not explicitly discourage such factual errors. To bridge this gap, we propose an evidence-grounded preference optimization based on reinforcement learning that explicitly teaches the model to prioritize genuine manipulation evidence over superficial reasoning patterns.

\subsubsection{Preference Dataset Construction}

We construct a preference dataset $\mathcal{D}_{\text{pref}}$ consisting of chosen-rejected explanation pairs. For each training sample with ground-truth explanation $e^*$ that accurately describes the specific manipulation traces present in the image, we  use advanced MLLM Qwen3.5-27B\cite{yang2025qwen3} to construct three types of rejected explanations based on the ground-truth: (1) \textit{Evidence-Omitted}: We instruct MLLM to remove key manipulation traces from $e^*$, producing an explanation that lacks certain critical evidence. (2) \textit{Irrelevant-induced}: We inject content that exists within image but is irrelevant to the deepfake judgement into $e^*$, simulating hallucinated evidence. (3) \textit{Combined above:} We further construct rejected explanations that not only lack critical evidence, but also injected with redundant information.  

For each sample with the $e^*$ as the chosen explanation, we pair a rejected explanation $e^-$, forming preference pairs $(e^+, e^-)$ where $e^+$ indicates the preferred explanation.
Then using $\mathcal{D}_{\text{pref}}$, we further fine-tune the SFT model using Direct Preference Optimization (DPO)~\cite{rafailov2023dpo}, which optimizes the policy $\pi_\theta$ to favor chosen explanations over rejected ones. Given a prompt $q$ (sample image with a prompt asking for detection rationale), the DPO loss is:

\begin{equation}
\scriptsize
\mathcal{L}_{\text{DPO}} = -\mathbb{E}_{(q, e^+, e^-) \sim \mathcal{D}_{\text{pref}}} \left[ \log \sigma\left( \beta \log\frac{\pi_\theta(e^+ \mid q)}{\pi_{\text{ref}}(e^+ \mid q)} - \beta \log\frac{\pi_\theta(e^- \mid q)}{\pi_{\text{ref}}(e^- \mid q)} \right) \right]
\end{equation}
where $\pi_{\text{ref}}$ is a reference model (the SFT checkpoint), and $\beta$ controls the deviation from the reference. This objective implicitly rewards explanations that are factually complete and penalizes those that omit evidence or include hallucinations, effectively aligning the model's reasoning with evidence-grounded  manipulation traces.


\subsubsection{Concise Explanation Model}

While detailed evidence explanations are valuable for forensic analysis, many real-world scenarios, such as social media fact-checking or real-time alerts, require brief, digestible outputs. To accommodate such use cases, we additionally develop a concise explanation model that describes the most critical manipulation cues into compact statements. This model is optimized via Group Relative Policy Optimization (GRPO)~\cite{shao2024grpo}, with reward functions regarding: (1) \textit{semantic fidelity}, which adopts BERTScore\cite{zhang2020bertscore} to measure  the semantic similarity between the rollout and reference explanation; and (2) \textit{conciseness}, which adopts SLEscore\cite{cripwell-etal-2023-simplicity} to penalize verbosity. The GRPO training encourages the model to find an optimal balance between overall reward, producing explanations that are both faithful and concise. 

\section{Dataset Overview.} The challenge uses XPlainVerse dataset\cite{narang2026xplainversemillionscalebenchmarkexplainable}, which is sourced from MultifakeVerse\cite{gupta2025multiversedeepfakesmultifakeversedataset} dataset. XPlainVerse is the largest Explainable Deepfake Detection Dataset containing 1 million images, including both authentic and manipulated images paired with reference explanations that ground the evaluation of model-generated justifications. This challenge contained a subset of XPlainVerse, separately contains 450k, 110k, 200k data in train, validation and test set.

\section{Experiments}

\noindent\textbf{Evaluation Metrics.} 
Deepfake detection is evaluated by detection accuracy and macro F1 score. Complex explanations are evaluated by BERTscore\cite{kenton2019bert}, along with Entity and Claim F1 that measures the semantic faithfulness in explanations. Simple explanations are evaluated with BERTscore along with SLE score\cite{cripwell-etal-2023-simplicity}, which measures the simplicity of prediction. The overall score is the weighted sum of all metrics. Refer to Challenge paper\cite{narang2026explainabledeepfakedetectionchallenge} for more metric details.

\subsection{Our experimental setup}

For the detection, we employ DINOv3~\cite{siméoni2025dinov3} as the visual encoder, fine-tuned via LoRA~\cite{hu2021lora}(rank=16, $\alpha$=16). We train it for 10 epochs with a batch size of 64 and an initial learning rate of $1\times10^{-4}$. $\lambda_1$/$\lambda_2$ is 0.5/0.2. $\tau$ is 0.2, $\beta$ is 0.5, EMA rate is 0.99. For explanation model, we adopt Qwen3-VL-8B-Instruct~\cite{yang2025qwen3} as base model and fine-tuned with LoRA (rank=32, $\alpha$=32).  The model first undergoes supervised fine-tuning on the ground-truth explanation annotations for 3 epochs, followed by reinforcement learning with DPO or GRPO algorithm for the evidence-grounded and concise explanation models, respectively. All experiments are conducted on 8 HUAWEI Ascend 910B NPUs.

\begin{table}[!th]
\centering
\caption{Public leaderboard Results of Top-5 Teams. Qwen3-VL-8B/InternVL3.5-13B are challenge\cite{narang2026explainabledeepfakedetectionchallenge} baselines.}
\vspace{-3mm}
\label{tab:main_results}
\resizebox{0.47\textwidth}{!}{
\begin{tabular}{@{}l|cccccccc@{}}
\toprule
\thead{Team\\name} & \thead{Overall\\ Score} & \thead{Detection\\ F1} & \thead{Complex\\ BERT} & \thead{Simple\\ Overall} & \thead{Explanation\\ Score} & \thead{Rank} \\
\midrule
\textbf{Pixel Slueth(Ours)} & \textbf{0.841617}&	0.942399&\textbf{0.706168}&\textbf{0.775501}&\textbf{0.740834}&\textbf{1} \\
Team Antvengers&0.838154&\textbf{0.947941}&0.696179&	0.760555&0.728367&2 \\
MSUteam&0.8312&0.933982&0.70113&0.755707&0.728419 &3\\
HIT VIRLAB&0.822242&0.926326&0.698713&0.737602&0.718157 &4\\
Team1&0.814503&0.916683&0.696807&0.727839&0.712323& 5\\
\midrule
Qwen3-VL-8B\cite{narang2026explainabledeepfakedetectionchallenge} & 0.642776&	0.634234&	0.664774&	0.637863&	0.651318& -\\
InternVL3.5-13B\cite{narang2026explainabledeepfakedetectionchallenge}&0.637566	&0.625662	&	0.661781&	0.637161&	0.649471 & - \\
\bottomrule
\end{tabular}}
\end{table}

\begin{table}[!htbp]
\centering
\footnotesize
\caption{Final hidden-test results for Top-5 team.}
\label{tab:final_results}
\vspace{-3mm}
\resizebox{0.48\textwidth}{!}{
\begin{tabular}{@{}lccccccccc@{}}
\toprule
Team &
\thead{Detection \\ F1} &
\thead{Complex \\Bert} &
\thead{Simple \\Bert} &
\thead{SLE \\Score} &
\thead{Entity\\F1} &
\thead{Claim\\F1} &
\thead{Explain\\Score} &
\thead{Overall\\ Score} & Rank\\
\midrule
\textbf{\thead{Pixel Sleuth(Ours)}} &
0.9424 & \textbf{0.7063} & \textbf{0.6819} & \textbf{0.9782} & \textbf{0.5280} & \textbf{0.4205} & \textbf{0.5800} & \textbf{0.7612} & \textbf{1}\\
Antvengers &
\textbf{0.9479} & 0.6958 & 0.6699 & 0.9720 & 0.5078 & 0.3940 & 0.5618 & 0.7549 &2\\
MSUteam &
0.9340 & 0.7004 & 0.6509 & 0.9726 & 0.4987 & 0.3932 & 0.5571 & 0.7456 &3\\
Team1 &
0.9167 & 0.6959 & 0.6321 & 0.9335 & 0.5155 & 0.4021 & 0.5590 & 0.7378 &4\\
HIT VIRLAB &
0.9263 & 0.6988 & 0.6493 & 0.9287 & 0.4467 & 0.3436 & 0.5235 & 0.7249 &5\\
\bottomrule
\end{tabular}}
\vspace{-0.75em}
\end{table}

\begin{figure*}[!t]
    \centering
    \includegraphics[width=0.9\linewidth]{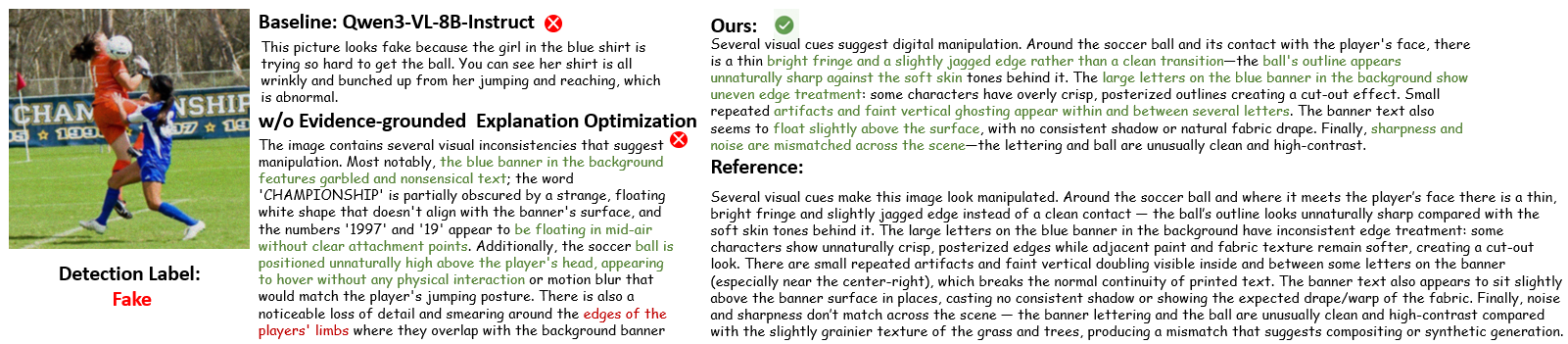}
    \vspace{-4mm}
    \caption{Visualizations for deepfake explanation. Result of baseline\cite{yang2025qwen3} shows irrelevant information, fails to identify valid evidence. SFT-only model captures partial evidence (marked in green) but still suffers irrelevant details(marked in red). Our model generates complete and accurate explanation(marked by green) without introducing spurious information. }
    \label{fig:vis_explain}
\end{figure*}
\begin{figure}[!h]
    \centering
    \includegraphics[width=0.95\linewidth]{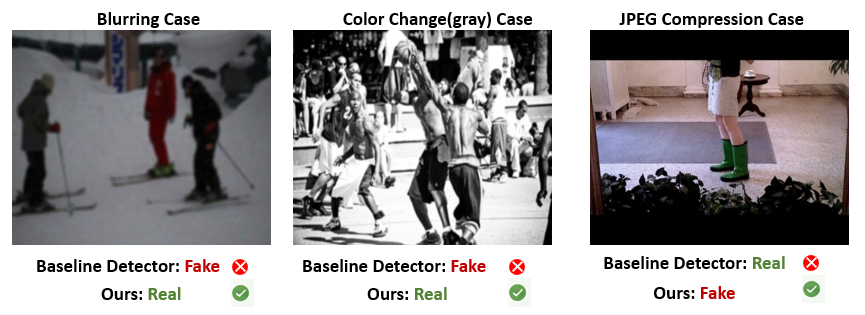}
    \vspace{-4mm}
    \caption{Qualitative comparison of detection robustness under samples with various image degradations. The baseline model misclassifies degraded samples, while our model shows correct predictions across different degradation types.}
    \label{fig:robust}
\end{figure}
\subsection{Challenge Results}
We present the public leaderboard results\footnote{The challenge website is \href{https://explainable-deepfake-detection.github.io}{https://explainable-deepfake-detection.github.io}, and Public leaderboard is shown in \href{https://www.codabench.org/competitions/16461/\#/results-tab}{https://www.codabench.org/competitions/16461/\#/results-tab}.} on the Challenge test set in Table~\ref{tab:main_results}, and the final hidden-test results(where Entity and Claim F1 Score are reported on a 10k hidden evaluation subset) in Table.\ref{tab:final_results}, comparing with other top-tier teams. Our method ranks first for the overall score among 138 participating teams with 414 submission entries in total. We yield the highest score for all explanation metrics, showcasing that we own (1) most accurate complex explanation, thanks to our evidence-grounded learning to guide the model to focus on the accurate forensic traces for more accurate and complete reasoning; (2) most concise and semantic correct simple explanation, which validates that our curated rewards can find an optimal balance between semantic fidelity and expression conciseness. As a result, our method yields the best overall score, showing detection robustness, explanation interpretability.


\vspace{-2mm}
\subsection{Qualitative Results.} 
\noindent\textbf{Robustness on Deepfake Detection.}
As shown in Figure~\ref{fig:robust}, the baseline model exhibits error predictions when confronted with quality-degraded samples. In contrast, our model maintains accurate predictions across different degradation types. The degradation-aware contrastive learning exposes the model to diverse degradation patterns during training, while the mean-teacher architecture ensures that augmented features are aligned toward a stable feature anchor, preventing feature drift. As a result, our model learns quality-invariant representations for deepfake detection.

\noindent\textbf{Evidence-accuracy on Explanation.}
As shown in Figure~\ref{fig:vis_explain}, the baseline's explanation is dominated by irrelevant information, indicating that without proper guidance, the MLLM tends to generate superficially plausible but factually ungrounded rationales. The SFT-only model, while benefiting from ground-truth supervision, achieves only partial correctness, as it correctly identifies some manipulation traces (e.g., unnatural text on the banner and floating contact) but still omits critical evidence such as the unnatural high-contrast of ball, and occasionally hallucinates non-existent artifacts like loss of detail on player's limbs.  This confirms our empirical finding that SFT alone is insufficient to ensure complete and faithful evidence grounding. In contrast, our model produces a concise yet comprehensive explanation that accurately enumerates all detectable manipulation traces without fabricating spurious details, demonstrating the effectiveness of evidence-grounded preference optimization in aligning model reasoning with factual evidence.

\subsection{Ablation Studies}
Since the ground-truth annotations for the Challenge test set are not publicly available, we instead conduct ablation studies on the training and validation sets, splitting them into 420k training samples and 140K testing samples to systematically validate the effectiveness of our designs.

\noindent\textbf{Ablation on Detection Designs.}
We choose the model of naive Dinov3 with MLP head as baseline and incorporate our designs on top of it. Results are shown in Table~\ref{tab:det_ablation}: (1)Degradation-aware Augmentation(``DA") brings a substantial improvement, confirming it effectively enhances its robustness to real-world quality variations. (2)Introducing supervised contrastive loss(``SC") yields further gain by pulling real and fake samples apart while aligning degraded versions of the same sample. (3)Incorporating mean-teacher stabilization(``MT") further boosts performance by anchoring augmented representations toward a stable feature reference to prevent potential feature drift caused by varied augmentations and ensures training stability.  Finally, the combination of above designs yields SOTA performance.

\begin{table}[!h]
\centering
\caption{Ablation on detection components.}
\label{tab:det_ablation}
\vspace{-3mm}
\resizebox{0.43\textwidth}{!}{
\begin{tabular}{@{}l|cc@{}}
\toprule
\textbf{Configuration} & \textbf{Detection F1} & \textbf{Detection Accuracy }\\
\midrule
Baseline & 0.8824 & 0.8877 \\
+ DA & 0.9402 &  0.9411\\
+ DA\& SC & 0.9588 & 0.9623 \\
+ DA\& MT & 0.9593 & 0.9627 \\
Full(+ DA\& MT \& SC) & \textbf{0.9724} & \textbf{0.9793} \\
\bottomrule
\end{tabular}}
\end{table}

\begin{table}[!h]
\centering
\caption{Ablation on complex explanation model designs.}
\label{tab:exp_ablation}
\vspace{-3mm}
\resizebox{0.43\textwidth}{!}{
\begin{tabular}{@{}l|ccc@{}}
\toprule
\textbf{Training Strategy} & \textbf{Fact F1} & \textbf{Claim F1}& \textbf{BERTScore} \\
\midrule
Baseline &0.4402 &0.3362 & 0.6038 \\
+ SFT & 0.5173  &0.4218 & 0.6673\\
+ DPO (Omit Only) & 0.5666 & 0.4860 &0.6912 \\
+ DPO (Irrelevant Only) & 0.5743 & 0.5042& 0.6856\\
+ DPO (Irrelevant + Omit )& 0.5942 & 0.5144 & 0.7131\\
Full(DPO with all three types) & \textbf{0.5969} & \textbf{0.5227} &\textbf{0.7184} \\
\bottomrule
\end{tabular}}
\end{table}
\noindent\textbf{Ablation on Deepfake Explanation Designs.}
We use Qwen3-VL-8B-Instruct\cite{yang2025qwen3} as baseline, and incorporate different training strategies on top for evidence-grounded explanation. Since the official evaluation code for Fact and Claim F1 takes a considerable amount of time, we randomly sample a 1k subset from our test set for evaluation. Table~\ref{tab:exp_ablation} reports the performance. (1) SFT improves all metrics by enabling the model to learn the basic pattern of evidence description. (2) Applying DPO with evidence-omitted rejected responses further improves performance by explicitly teaching the model to avoid missing critical manipulation traces, and using irrelevant-augmented samples enhances the model's ability to suppress hallucinated details. (3) Combining all three types(irrelevant-induced, evidence-omitted and combined above) of rejected samples yields the best results, confirming that evidence omission and irrelevant hallucination are complementary failures we addressed.


\section{Conclusion}
\label{sec:conclusion}

In this work, we propose a framework for the explainable deepfake detection task.
For robust detection, we introduced degradation-aware contrastive learning with a mean-teacher architecture that prevents feature drift while preserving augmentation benefits. For faithful explanation, we devised evidence-grounded preference optimization via DPO that explicitly teaches the model to prioritize genuine manipulation evidence, complemented by a concise explanation model optimized through GRPO with tailored reward for brevity-sensitive scenarios. Our method ranks first for ACM
Multimedia 2026 Explainable Deepfake Detection Challenge, providing a promising method for future research toward more reliable and transparent forensic AI systems.

\noindent\textbf{Acknowledgements.}This work was supported by the grants from the National Natural Science Foundation of China (62372014, 62525201,
62132001, 62432001), Beijing Nova Program and Beijing
Natural Science Foundation (4252040, L247006).
\bibliographystyle{ACM-Reference-Format}
\bibliography{references}

\end{document}